\documentclass[10pt,twocolumn,letterpaper]{article}

\usepackage[pagenumbers]{cvpr} 

\usepackage{graphicx}
\usepackage{amsmath}
\usepackage{amssymb}
\usepackage{booktabs}
\usepackage[square,sort,comma,numbers]{natbib}
\usepackage{multirow}
\usepackage{arydshln}
\usepackage{float}
\setcitestyle{square}
\setcitestyle{comma}

\usepackage[pagebackref,breaklinks,colorlinks]{hyperref}

\usepackage[capitalize]{cleveref}
\crefname{section}{Sec.}{Secs.}
\Crefname{section}{Section}{Sections}
\Crefname{table}{Table}{Tables}
\crefname{table}{Tab.}{Tabs.}

\def\cvprPaperID{27} 
\def\confName{CV4Animals}
\def\confYear{2024}

\begin{document}

\title{\vspace{-25pt}ChimpVLM: Ethogram-Enhanced Chimpanzee Behaviour Recognition}\vspace{-10pt}

\author{
Otto Brookes\\
University of Bristol\\
Bristol, United Kingdom\\
{\tt\small otto.brookes@bristol.ac.uk}
\and
Majid Mirmehdi\\
University of Bristol\\
Bristol, United Kingdom\\
{\tt\small majid@cs.bris.ac.uk}
\and
Hjalmar Kühl\\
iDiv\\
Leipzig, Germany\\
{\tt\small hjalmar.kuehl@idiv.de}
\and
Tilo Burghardt\\
University of Bristol\\
Bristol, United Kingdom\\
{\tt\small tilo@cs.bris.ac.uk}
}

\maketitle

\begin{abstract}
We show that chimpanzee behaviour understanding from camera traps can be enhanced by providing visual architectures with access to an embedding of text descriptions that detail species behaviours. In particular, we present a vision-language model which employs multi-modal decoding of visual features extracted directly from camera trap videos to process query tokens representing behaviours and output class predictions. Query tokens are initialised using a standardised ethogram of chimpanzee behaviour, rather than using random or name-based initialisations. In addition, the effect of initialising query tokens using a masked language model fine-tuned on a text corpus of known behavioural patterns is explored. We evaluate our system on the PanAf500 and PanAf20K datasets and demonstrate the performance benefits of our multi-modal decoding approach and query initialisation strategy on multi-class and multi-label recognition tasks, respectively. Results and ablations corroborate performance improvements. We achieve state-of-the-art performance over vision and vision-language models in top-1 accuracy (+6.34\%) on PanAf500 and overall (+1.1\%) and tail-class (+2.26\%) mean average precision on PanAf20K. We share complete source code and network weights for full reproducibility of results and easy utilisation.
\end{abstract}\vspace{-18pt}

\section{Introduction}\vspace{-5pt}
\label{sec:intro}

As the climate crisis deepens, the threat to many endangered species grows ever more perilous~\cite{almond2022living}. For example, all great ape species are now classified as endangered or critically endangered~\cite{IUCN}. Consequently, there is urgent need for methods that can help to monitor populations and assess the effectiveness of conservation interventions~\cite{kuhl2013animal,congdon2022future,tuia2022}. These include the monitoring of species behaviours, on which we focus here, noting that they can provide some of the earliest indicators of population stress~\cite{dominoni2020conservation,carvalho2022using,chappell2022role}.

\begin{figure}[!h]
\centering
\includegraphics[width=\linewidth, height=105pt]{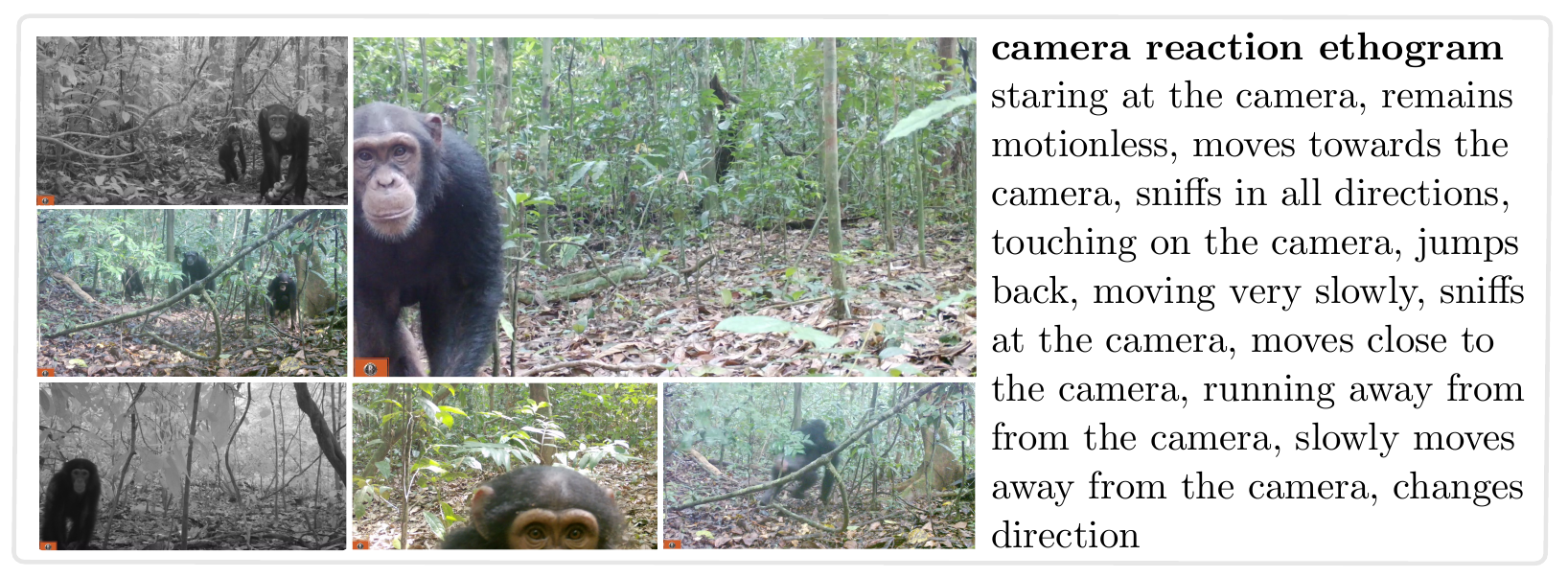}\vspace{-10pt}
\caption{\textbf{Camera Reaction Ethogram}. A series of still frames (left) demonstrates the intra-class variation of the camera reactivity behaviour. The ethogram (right) describes the behavioural patterns corresponding to each of the different variations and provides a semantically richer representation of the behaviour.}\vspace{-10pt} 
\label{fig:dataset_overview}
\end{figure}

Most behavioural data on great apes is collected from camera traps~\citep{burton2015wildlife, steenweg2017scaling} and footage is highly diverse (see Fig.~\ref{fig:dataset_overview}). Great apes display enormous behavioural variation across countries, populations and habitats~\citep{kalan2020environmental} and footage is dominated by commonly occurring behaviours leading to a comparative lack of imagery for rare classes~\cite{brookes2024panaf20k}. Consequently, a reliable automatic classification of complex behaviours as performed by chimpanzees in camera trap footage, is challenging.

Several works have combined visual and audio modalities~\cite{sakib2020visual, bain2021automated, brookes2024panaf20k} for enhancing the classification of behaviour. However, performance often falls short of what is required for truly meaningful ecological analyses, even when enhanced with long-tail learning techniques~\citep{brookes2023triple}. Although domain knowledge of great ape behaviour is readily encoded by biologists in text form via ethograms~\citep{nishida1999ethogram, nishida2010chimpanzee, zamma_matsusaka_2015}, they have remained largely unexplored for aiding classification of behaviours. Therefore, here we propose integrating a language model to improve classification results, since it can effectively represent semantics and abstract away visual variance~\citep{el2023learning}. Additionally, vision language models (VLMs) have recently achieved SOTA results on benchmark human action recognition datasets, demonstrating outstanding performance on few-shot recognition tasks~\citep{yu2022coca, xue2022clip, zhao2023learning} which are analogous to long-tail learning problems~\citep{perrett2023}.

We approach the classification task using a decoder-based architecture, using visual features extracted directly from camera traps to decode query tokens representing great ape behaviours. Query tokens are initialised by embedding ethogram information using a behaviour-specific language model that captures fine-grained, domain-specific semantics. As summarised in Fig~\ref{fig:t2v_decoder}, our approach, is simple, lightweight, and achieves state-of-the-art (SOTA) results on the largest public dataset applicable to the task~\citep{brookes2024panaf20k}.

\section{Related Work}\vspace{-5pt}
\textbf{Great Ape Action \& Behaviour Recognition}. The first automated great ape behaviour recognition system~\cite{sakib2020visual} for camera trap videos employed a two-stream architecture~\cite{simonyan2014two} that exploited both RGB and optical flow for classification. While a strong top-1 performance at 73.52\% was reported, the authors acknowledged relatively low average per class accuracy at 42.33\%. Subsequent work introduced an audio stream to aid with classification~\citep{bain2021automated}, reporting high average precision on two audio-visual behaviours. More recently, a triple-stream architecture that examines the role of pose and long-tail recognition techniques was introduced~\cite{brookes2023triple}. As far as we are aware, to date no work has explored language as a modality for improving behaviour classification.

\textbf{Vision-Language Models}. VLMs have attracted significant attention owing to their impressive performance on a variety of computer vision tasks~\cite{zhang2024vision}. Driven by the contrastive image-text pre-training regime established by CLIP~\citep{radford2021learning}, vision models are guided to produce highly generalisable representations~\citep{yu2022coca}. Several systems have successfully extended this paradigm to spatio-temporal data and have demonstrated SOTA performance on tasks requiring strong inductive biases, such as zero~\citep{lei2021less} and few-shot recognition~\citep{xue2022clip}. Alternatively, a lightweight decoder architecture may comprise several cross-attention layers to promote close correspondence between modalities~\citep{lin2022frozen, zhao2023learning}. In this work, we extend this base architecture with a query token initialisation strategy that exploits ethograms for enhancing behaviour recognition directly.
\vspace{-5pt}
\section{Method}\vspace{-5pt}

\begin{figure}[!h]
\centering
\includegraphics[width=235pt,height=205pt]{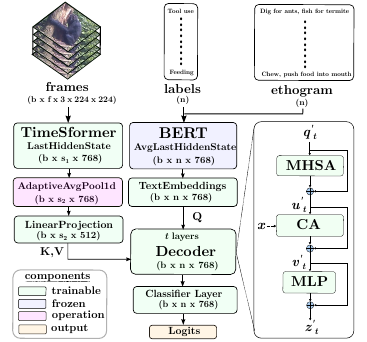}
\caption{{\textmd{\textbf{System Overview}. Our proposed system employs a standard vision encoder, a fine-tuned large language model, and a multi-model decoder comprising several cross-attention layers. Query tokens are initialised by embedding either behaviour names or descriptions using a large language model. During training, the multi-modal decoder utilises spatio-temporal visual features to decode query tokens and output logits for downstream classification.\vspace{-10pt}}}}
\label{fig:t2v_decoder}
\end{figure}

\textbf{Overview}. Our system comprises a standard vision model, a masked language model, and a multi-modal decoder (see Fig~\ref{fig:t2v_decoder}). First,  BERT~\citep{devlin2018bert} is fine-tuned on a comprehensive description of \textit{all} known behaviour patterns of chimpanzees~\citep{nishida2010chimpanzee}. Prior to model training, BERT was used to initialise learnable query tokens by embedding either behaviour names (equivalent to their respective classes) or descriptions extracted from an ensemble of behavioural ethograms. The multi-model decoder, which comprises several cross-attention layers, then utilises spatio-temporal features extracted by the vision model to decode the query tokens and output logits. Note that the embedding of ethogram descriptions is built  once for a species and can be reused for training on different visual datasets.

\textbf{Video Encoding}. The vision backbone processes an input video $V \in \mathbb{R}^{T \times H \times W \times 3}$ to produce a feature representation  $x \in \mathbb{R}^{T \times D}$ of $V$ where $T$, $H$, $W$ and $D$ represent the temporal, height, width, and embedding dimension, respectively. A 1D adaptive pooling operation is applied to the output feature $x$ to ensure $T$ is consistent with the input video. In our system, the video features $x$ act as keys in the cross-attention layers of our decoder module.

\textbf{Query Generation}. As described, a behaviour-specific masked language model, BERT~\cite{devlin2018bert}, is used is to initialise query tokens. This is realised by generating text embeddings from two sources of text: behaviour \textit{names} and \textit{descriptions}~\citep{nishida2010chimpanzee}. The descriptions are extracted from ethograms utilised in field studies on great apes and provide a more detailed account of each behaviour. Behaviour names simply reference their class label. These embeddings serve as per-behaviour query tokens $q \in \mathbb{R}^{C \times D }$ where $C$ is the total number of behaviours and $D$ is the embedding dimension. The final query tokens are instantiated as a learnable parameter at the beginning of training while the language model remains frozen throughout.

\textbf{Multi-modal Decoding}. A multi-modal decoder utilises the video features $x$ to decode the queries $q$. Each decoder layer comprises multi-headed self-attention (MHSA), cross attention (CA), and a multi-layer perceptron (MLP). The decoder layer performs self-attention on the text queries, cross-attention between the queries and the outputs of the spatio-temporal video encoder, followed by the application of a multilayer perceptron:
\begin{align}
    u'_t &= \text{MHSA}(q'_t) + q'_t, \quad \\
    v'_t &= \text{CA}(u'_t, x) + u'_t, \quad \\
    z'_t &= \text{MLP}(v'_t) + v'_t,  \quad \\
\end{align}
where $u'_t$ and $v'_t$ describe intermediate representations from the \text{MHSA} and \text{CA} operations, respectively, and $z'_t$ is the output of the $t^{th}$ decoder layer.

\textbf{Classification}. Finally, a classifier $W \in \mathbb{R}^{C\times D}$ is used to process the output of the decoder and generate logits where $C$ represents the total number of behaviours. Optimisation is performed via the standard softmax cross-entropy and binary cross-entropy losses for the multi-class and multi-label setting, respectively.
\vspace{-5pt}
\section{Experiments and Results}
\vspace{-5pt}
\subsection{Setup}
\vspace{-4pt}
\textbf{Training Details}. We trained all \textit{vision} models with AdamW optimisation using a batch size of 64 and performed linear warm-up followed by cosine annealing using an initial learning rate of $1\times10^{-5}$ that increases to $1\times10^{-4}$ over 10 epochs. Feature extractor backbones were initialised with Kinetics-400~\cite{kay2017kinetics} pre-trained weights. Similarly, we trained all \textit{language} models with AdamW optimisation using a batch size of 128, learning rate of $2\times10^{-5}$, weight decay of $0.01$ and a masking proportion of $m={0.2}$. All training runs used 8 Tesla V100 GPUs stopping at 100 epochs after validation performance had plateaued.

\textbf{Datasets}. We ran experiments on both parts of the PanAf20K dataset~\citep{brookes2024panaf20k} benchmarking multi-class and multi-label behaviour recognition on PanAf500 (see Sec.~\ref{subsec:p500}) and the full PanAf20k (see Sec~\ref{subsec:p20k}) data, respectively.

\textbf{Baseline Models}. We trained a CLIP-based model for comparison with our own system. To do this, we utilised Internvideo~\citep{wang2022internvideo} and concatenated video and text features during training, rather than using only vision, as in~\citep{wang2024exploring}. We also trained three popular vision-only models; 3D ResNet-50~\citep{du2017closer}, MViTv2~\citep{li2022mvitv2} and  Timesformer~\citep{bertasius2021space} architectures providing a direct vision-only comparison to our model.\vspace{-5pt}

\subsection{PanAf500}\label{subsec:p500}
\vspace{-5pt}
\textbf{Details}. We trained models as described in~\citep{brookes2024panaf20k}. During training and testing, we imposed a temporal threshold that only frame sequences in which a behaviour is exhibited for at least $t=16$ consecutive frames are utilised. This is to ensure that only well-defined behaviour instances were retained. We then sub-sampled 16-frame sequences from clips that satisfy the behaviour threshold. All behavioural action recognition models were evaluated using average top-1 and average per-class accuracy (C-Avg). Since the PanAf500 dataset comprises behavioural actions~\citep{brookes2023triple} rather than more complex behaviours, we initialise query tokens using only behaviour names~(CLS). Models utilising a pretrained and fine-tuned language model are denoted with +PT and +FT, respectively.

\begin{figure*}[!h]
\begin{minipage}{0.48\textwidth}\vspace{-15pt}
    \includegraphics[width=\linewidth]{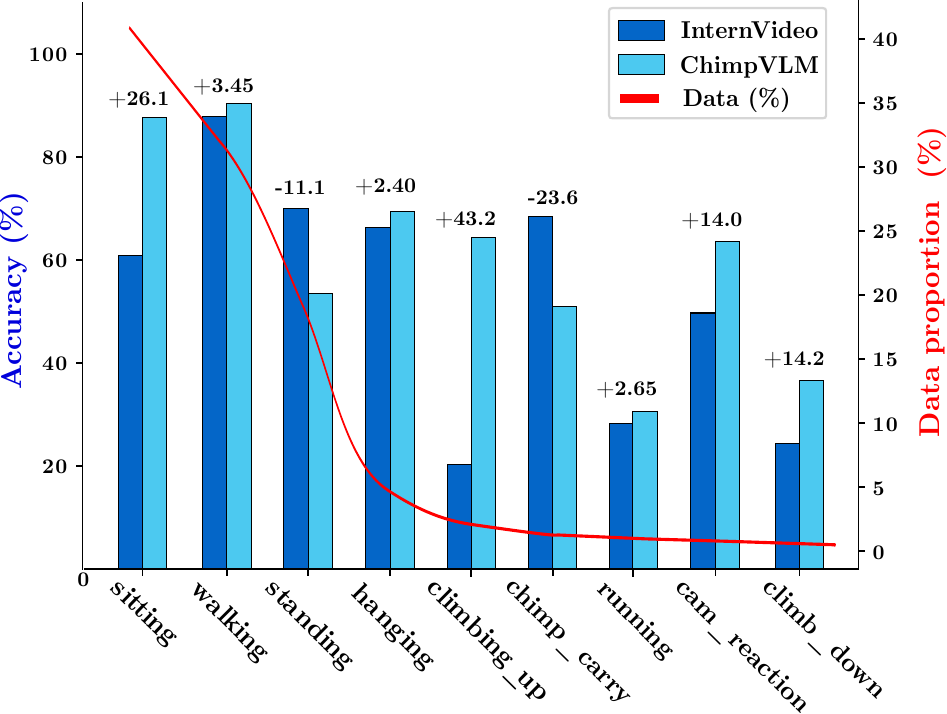}\vspace{-7pt}
    \caption{\textbf{PanAf500 Class-wise Performance vs. Proportion of Data}. The per-class accuracy of ChimpVLM (CLS+FT) and Internvideo is plotted against the proportion of data for each class in the PanAf500 dataset.\vspace{-10pt}}
    \label{fig:p500_per_class_performance}
    \end{minipage}
    \hspace{\fill} 
    \begin{minipage}{0.48\textwidth}\vspace{-15pt}
    \includegraphics[width=\linewidth]{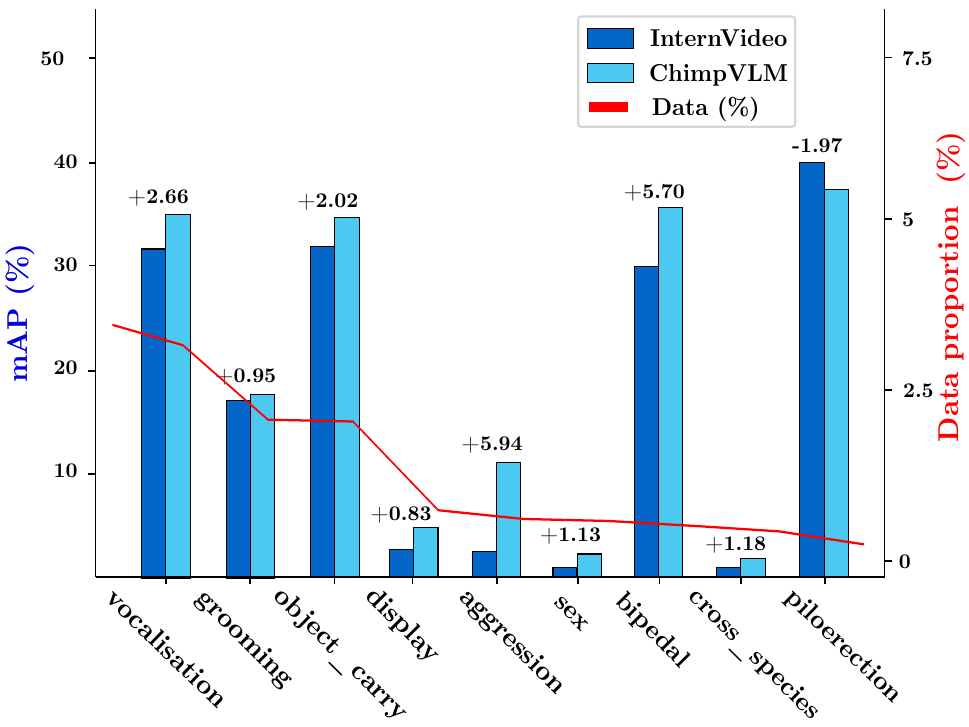}\vspace{-7pt}
    \caption{\textbf{PanAf20K Class-wise Performance vs. Proportion of Data}. The per-class AP of ChimpVLM (DSC+FT) and Internvideo is plotted against the proportion of data for each of the bottom 9 classes in the PanAf20k dataset.\vspace{-10pt}}
    \label{fig:p20k_per_class_performance}
    \end{minipage}
\end{figure*}

\begin{table}[!h]
\footnotesize
\centering
\caption{\textbf{PanAf500 Behavioural Action Recognition Benchmarks.} \textmd{Results are reported for the baseline 3D ResNet-50~\cite{hara2017learning}, MViTV2~\cite{li2021mvitv2}, TimeSformer~\cite{bertasius2021space}, Internvideo~\citep{wang2022internvideo} and ChimpVLM. The highest scores for top-1 and average per-class accuracy are shown in bold.}}
\begin{tabular}{l l c c c}
\toprule
 & \multirow{2}{*}{\textbf{Model}} & \multirow{2}{*}{\textbf{Params (M)}} & \multicolumn{2}{c}{\textbf{Accuracy (\%)}} \\
\cmidrule(lr){4-5}
 & & & \multicolumn{1}{c}{Top-1} & \multicolumn{1}{c}{C-Avg} \\
\midrule
1 & \multicolumn{1}{l}{3D R50~\cite{hara2017learning}} & 34.3 & 77.45 & 55.17 \\
2 & \multicolumn{1}{l}{MViTV2~\cite{li2021mvitv2}} & 36.4 & 78.31 & 40.45 \\
3 & \multicolumn{1}{l}{TimeSformer~\cite{bertasius2021space}} & 121.0 & 78.53 & 45.05 \\
4 & \multicolumn{1}{l}{Internvideo~\citep{wang2022internvideo}} & 478.0 & 78.57 & 54.01 \\
\midrule
 & \multicolumn{1}{l}{\textbf{ChimpVLM (ours)}} & & & \\
5 & \multicolumn{1}{l}{-- CLS+PT} & 167.0 & 81.36 & 57.56 \\
6 & \multicolumn{1}{l}{-- CLS+FT} & 167.0 & \textbf{84.91} & \textbf{61.94}\\
\bottomrule
\end{tabular}
\label{tab:behaviour_rec_results}
\end{table}

\textbf{Baselines}. We first establish baseline performance (rows 1-4 of Table~\ref{tab:behaviour_rec_results}). The best top-1 performance is achieved by the VLM, Internvideo, although it only marginally improves on the vision-only models (+0.03\%). While the top-1 accuracy of MViT2 and TimeSformer are marginally better than the 3D ResNet-50, they are significantly outperformed with respect to average per-class accuracy. Internvideo is also outperformed by TimeSformer, albeit to a lesser extent~(-1.16\%).

\textbf{Multi-Modal Decoding \& Language Model Fine-tuning}. As shown in Table~\ref{tab:behaviour_rec_results} for PanAf500, models using the multi-modal decoding strategy outperform the baseline models across both metrics (rows 5-6). The model with query tokens initialised using the fine-tuned masked language model, ChimpVLM~(CLS+FT), outperforms all other models significantly. With respect to top-1 and average per-class accuracy, ChimpVLM (CLS+FT) achieves performance increases of 6.34\% and 7.93\% in top-1 and average per-class accuracy over the closest competing model (i.e., Internvideo). As shown in Fig~\ref{fig:p500_per_class_performance}, performance improvements can be observed across the majority of classes and particularly at the tail end of the distribution. Per-class accuracy increases of 14.2\%, 14.0\% and 2.65\% are achieved on the three rarest classes (i.e., running, camera reaction and climbing down). Although models using query tokens initialised with the language models pre-trained weights, ChimpVLM (CLS+PT), still outperform the baseline models, the performance gains are reduced (+2.79\% and +3.55\% over Internvideo for top-1 and average per-class accuracy, respectively). Thus, our results suggest that in-domain fine-tuning of the language model is indeed beneficial for our multi-modal decoding approach when deployed in a multi-class setting.
\vspace{-5pt}
\subsection{PanAf20k}\label{subsec:p20k}
\vspace{-5pt}
\textbf{Details}. As per~\citep{brookes2024panaf20k}, we uniformly sub-sampled 16 frames from each video during training. All models were evaluated using macro-averaged mean average precision (mAP). Behaviour classes were grouped, based on class frequency, into head~($>10\%$), middle~($>1\%$), and tail~($<1\%$) segments, and mAP performance is reported for each segment in Table~\ref{tab:multilabel_benchmarks}.

\begin{table}[!h]
\begin{center}
\caption{\textbf{PanAf20K Multi-label Behaviour Recognition Benchmarks.} Results are reported as segment mAPs for 3D ResNet-50~\citep{hara2017learning}, MViTV2~\citep{li2022mvitv2}, TimeSformer~\citep{bertasius2021space}, Internvideo~\citep{wang2022internvideo} and ChimpVLM. The highest score for each segment all is shown in bold.}
\resizebox{\columnwidth}{!}{%
\begin{tabular}{cccrrrr}
\toprule
 & \multirow{2}{*}{\textbf{Model}}
 & \multirow{2}{*}{\textbf{Params (M)}}
& \multicolumn{4}{c}{\textbf{mAP (\%)}} \\
\cmidrule(r){4-7}
 & & & \multicolumn{1}{c}{All} & \multicolumn{1}{c}{Head} & \multicolumn{1}{c}{Middle} & \multicolumn{1}{c}{Tail} \\
 \midrule
1 & \multicolumn{1}{l}{3D R50~\cite{hara2017learning}} & 34.3 & 46.03 & 86.12 & \textbf{53.22} & 9.73 \\
2 & \multicolumn{1}{l}{MViTV2~\cite{li2021mvitv2}} & 36.4 & 45.71 & 88.72 & 51.16 & 9.76 \\
3 & \multicolumn{1}{l}{TimeSformer~\cite{bertasius2021space}} & 121.0 & 47.24 & \textbf{88.83} & 51.91 & 13.29 \\
4 & \multicolumn{1}{l}{Internvideo (3-layer)} & 478.0 & 46.90 & 85.58 & 49.92 & 14.10 \\
 \midrule
& \textbf{ChimpVLM (ours)} & \multicolumn{1}{l}{} & & & \\
5 & \multicolumn{1}{l}{– CLS+PT} & 167.0 & 44.56 & 85.92 & 48.61 & 11.45 \\
6 & \multicolumn{1}{l}{– DSC+PT} & 167.0 & 45.73 & 86.53 & 51.35 & 15.11 \\
7 & \multicolumn{1}{l}{– CLS+FT} & 167.0 & 45.87 & 87.12 & 45.28 & 16.24 \\
8 & \multicolumn{1}{l}{– DSC+FT} & 167.0 & \textbf{48.48} & 87.91 & 52.95 & \textbf{16.36} \\
\bottomrule
\end{tabular}
}
\label{tab:multilabel_benchmarks}
\end{center}\vspace{-18pt}
\end{table}

\textbf{Baselines}. In contrast to the baseline results reported in Sec.~\ref{subsec:p20k}, the vision-only model, TimeSformer, outperforms Internvideo for this larger dataset across the majority of segments with the exception of the tail (see rows 3 and 4).

\textbf{Multi-Modal Decoding, Language Model Fine-tuning \& Query Initialisation.} As can be seen in Table~\ref{tab:multilabel_benchmarks}, the best overall performance is achieved by the model initialised with queries generated from full behaviour descriptions using the fine-tuned language model, ChimpVLM (DSC+FT). It achieves the best performance on overall and tail class mAP, improving upon the closest competing model by +1.1\% and +2.26\%, respectively. As shown in Fig.~\ref{fig:p20k_per_class_performance}, ChimpVLM (DSC+FT) achieves improved performance on the majority of tail classes, being outperformed on only one class (i.e., piloerection). Models initialised with queries generated from only the behaviour names (CLS) perform relatively poorly on overall mAP regardless of whether the language model is pretrained or fine-tuned (row 5 and 7), although strong improvements in average per-class performance are reported for ChimpVLM (CLS+FT). In general, we observe that models initialised with full behaviour descriptions outperform their class-only counterparts. This effect appears more pronounced for models that utilise the fine-tuned language model, where performance improvements of +2.61\% on overall mAP are reported (rows 7-8). Similarly, models that utilise query tokens initialised with the fine-tuned language model outperform their pretrained counterparts. The ChimpVLM models generally perform better on tail classes than their VLM and vision-only baselines. However, ChimpVLM (DSC+FT) is the only model, which achieves higher overall mAP and is still outperformed on head and middle classes, albeit marginally (by 0.92\% and 0.27\%, respectively). Our results suggest that language model fine-tuning and query initialisation using entire behaviour descriptions is required to fully realise the benefit of ethogram-based information utilisation.
\vspace{-5pt}
\section{Conclusion}\vspace{-5pt}
In this work, we showed that chimpanzee behaviour recognition can be enhanced by  embedding ethogram-based text descriptions that detail species behaviours. We demonstrated this by constructing a VLM, centred around a multi-modal decoder that is initialised via ethogram information and visually trained camera trap datasets. We showed that such a decoder-based architecture can achieve state-of-the-art performance, beyond that of standard CLIP-based architectures, on the multi-class and multi-label behaviour recognition tasks of the PanAf500 and PanAf20k datasets, respectively. We further demonstrated that initialising query tokens using text embeddings of behaviour descriptions, as opposed to the name of behaviours only, can benefit recognition performance. Finally, we provided evidence  that in-domain fine-tuning of language models can be an effective way to achieve further performance improvements. We hope that this work will inspire further exploration of fusing ethogram and visual information for the benefit of nature.

\section*{\large{Acknowledgements}\vspace{-5pt}}
\label{SecAck}
We thank the Pan African Programme: ‘The Cultured Chimpanzee’ team and its collaborators for allowing the use of their data for this paper. We thank Amelie Pettrich, Antonio Buzharevski, Eva Martinez Garcia, Ivana Kirchmair, Sebastian Schütte, Linda Gerlach and Fabina Haas. We also thank management and support staff across all sites; specifically Yasmin Moebius, Geoffrey Muhanguzi, Martha Robbins, Henk Eshuis, Sergio Marrocoli and John Hart. Thanks to the team at https://www.chimpandsee.org particularly Briana Harder, Anja Landsmann, Laura K. Lynn, Zuzana Macháčková, Heidi Pfund, Kristeena Sigler and Jane Widness. The work that allowed for the collection of the dataset was funded by the Max Planck Society, Max Planck Society Innovation Fund, and Heinz L. Krekeler. In this respect we would like to thank: Ministre des Eaux et Forêts, Ministère de l'Enseignement supérieur et de la Recherche scientifique in Côte d’Ivoire; Institut Congolais pour la Conservation de la Nature, Ministère de la Recherche Scientifique in Democratic Republic of Congo; Forestry Development Authority in Liberia; Direction Des Eaux Et Forêts, Chasses Et Conservation Des Sols in Senegal; Makerere University Biological Field Station, Uganda National Council for Science and Technology, Uganda Wildlife Authority, National Forestry Authority in Uganda; National Institute for Forestry Development and Protected Area Management, Ministry of Agriculture and Forests, Ministry of Fisheries and Environment in Equatorial Guinea. This work was supported by the UKRI CDT in Interactive AI under grant EP/S022937/1.

{\small
\bibliographystyle{ieee_fullname}
\bibliography{egbib}

\begin{thebibliography}{10}\itemsep=-1pt

\bibitem{almond2022living}
R.E.A. Almond, M. Grooten, D Juffe~Bignoli, and T Petersen.
\newblock Wwf (2022) living planet report 2022 - building a nature-positive society.
\newblock 2022.

\bibitem{bain2021automated}
Max Bain, Arsha Nagrani, Daniel Schofield, Sophie Berdugo, Joana Bessa, Jake Owen, Kimberley~J Hockings, Tetsuro Matsuzawa, Misato Hayashi, Dora Biro, et~al.
\newblock Automated audiovisual behavior recognition in wild primates.
\newblock {\em Science Advances}, 7(46):eabi4883, 2021.

\bibitem{bertasius2021space}
Gedas Bertasius, Heng Wang, and Lorenzo Torresani.
\newblock Is space-time attention all you need for video understanding?
\newblock In {\em Proceedings of the International Conference on Machine Learning (ICML)}, July 2021.

\bibitem{brookes2023triple}
Otto Brookes, Majid Mirmehdi, Hjalmar~S. K{\"{u}}hl, and Tilo Burghardt.
\newblock Triple-stream deep metric learning of great ape behavioural actions.
\newblock In {\em Proceedings of the 18th International Joint Conference on Computer Vision, Imaging and Computer Graphics Theory and Applications}, pages 294--302, 2023.

\bibitem{brookes2024panaf20k}
Otto Brookes, Majid Mirmehdi, Colleen Stephens, Maureen McCarthy, Mizuki Murai, Emmanuelle Normand, Virginie Vergnes, Amelia Meier, Juan Lapuente, Roman Wittig, Dervla Dowd, Sorrel Jones, Vera Leinert, Erin Wessling, Katherine Corogenes, Klaus Zuberb{\"u}hler, Kevin Lee, Samuel Angedakin, Kevin Langergraber, Paula Dieguez, Nuria Maldonado, Christophe Boesch, Mimi Arandjelovic, Hjalmar K{\"u}hl, and Tilo Burghardt.
\newblock Panaf20k: A large video dataset for wild ape detection \& behaviour analysis.
\newblock {\em International Journal of Computer Vision (IJCV)}, 2024.

\bibitem{burton2015wildlife}
A~Cole Burton, Eric Neilson, Dario Moreira, Andrew Ladle, Robin Steenweg, Jason~T Fisher, Erin Bayne, and Stan Boutin.
\newblock Wildlife camera trapping: a review and recommendations for linking surveys to ecological processes.
\newblock {\em Journal of applied ecology}, 52(3):675--685, 2015.

\bibitem{carvalho2022using}
Susana Carvalho, Erin~G Wessling, Ekwoge~E Abwe, Katarina Almeida-Warren, Mimi Arandjelovic, Christophe Boesch, Emmanuel Danquah, Mamadou~Saliou Diallo, Catherine Hobaiter, Kimberley Hockings, et~al.
\newblock Using nonhuman culture in conservation requires careful and concerted action.
\newblock {\em Conservation Letters}, 15(2):e12860, 2022.

\bibitem{chappell2022role}
Jackie Chappell and Susannah~KS Thorpe.
\newblock The role of great ape behavioral ecology in one health: Implications for captive welfare and re-habilitation success.
\newblock {\em American journal of primatology}, 84(4-5):e23328, 2022.

\bibitem{congdon2022future}
JV Congdon, M Hosseini, EF Gading, M Masousi, M Franke, and SE MacDonald.
\newblock The future of artificial intelligence in monitoring animal identification, health, and behaviour, 2022.

\bibitem{devlin2018bert}
Jacob Devlin, Ming-Wei Chang, Kenton Lee, and Kristina Toutanova.
\newblock Bert: Pre-training of deep bidirectional transformers for language understanding.
\newblock In {\em North American Chapter of the Association for Computational Linguistics}, 2019.

\bibitem{dominoni2020conservation}
Davide~M Dominoni, Wouter Halfwerk, Emily Baird, Rachel~T Buxton, Esteban Fern{\'a}ndez-Juricic, Kurt~M Fristrup, Megan~F McKenna, Daniel~J Mennitt, Elizabeth~K Perkin, Brett~M Seymoure, et~al.
\newblock Why conservation biology can benefit from sensory ecology.
\newblock {\em Nature Ecology \& Evolution}, 4(4):502--511, 2020.

\bibitem{el2023learning}
Mohamed El~Banani, Karan Desai, and Justin Johnson.
\newblock Learning visual representations via language-guided sampling.
\newblock In {\em Proceedings of the IEEE/CVF Conference on Computer Vision and Pattern Recognition}, pages 19208--19220, 2023.

\bibitem{hara2017learning}
Kensho Hara, Hirokatsu Kataoka, and Yutaka Satoh.
\newblock Learning spatio-temporal features with 3d residual networks for action recognition.
\newblock In {\em Proceedings of the IEEE International Conference on Computer Vision Workshops,}, pages 3154--3160, 2017.

\bibitem{IUCN}
IUCN.
\newblock Iucn red list of threatened species version 2022.1.
\newblock 2022.

\bibitem{kalan2020environmental}
Ammie~K Kalan, Lars Kulik, Mimi Arandjelovic, Christophe Boesch, Fabian Haas, Paula Dieguez, Christopher~D Barratt, Ekwoge~E Abwe, Anthony Agbor, Samuel Angedakin, et~al.
\newblock Environmental variability supports chimpanzee behavioural diversity.
\newblock {\em Nature Communications}, 11(1):4451, 2020.

\bibitem{kay2017kinetics}
Will Kay, Joao Carreira, Karen Simonyan, Brian Zhang, Chloe Hillier, Sudheendra Vijayanarasimhan, Fabio Viola, Tim Green, Trevor Back, Paul Natsev, et~al.
\newblock The kinetics human action video dataset.
\newblock {\em arXiv preprint arXiv:1705.06950}, 2017.

\bibitem{kuhl2013animal}
Hjalmar~S K{\"u}hl and Tilo Burghardt.
\newblock Animal biometrics: quantifying and detecting phenotypic appearance.
\newblock {\em TREE}, 28(7):432--441, 2013.

\bibitem{lei2021less}
Jie Lei, Linjie Li, Luowei Zhou, Zhe Gan, Tamara~L Berg, Mohit Bansal, and Jingjing Liu.
\newblock Less is more: Clipbert for video-and-language learning via sparse sampling.
\newblock In {\em Proceedings of the IEEE/CVF conference on computer vision and pattern recognition}, pages 7331--7341, 2021.

\bibitem{li2021mvitv2}
Yanghao Li, Chaoxia Wu, Haoqi Fan, Karttikeya Mangalam, Bo Xiong, Jitendra Malik, and Christoph Feichtenhofer.
\newblock Mvitv2: Improved multiscale vision transformers for classification and detection.
\newblock {\em 2022 IEEE/CVF Conference on Computer Vision and Pattern Recognition (CVPR)}, pages 4794--4804, 2021.

\bibitem{li2022mvitv2}
Yanghao Li, Chao-Yuan Wu, Haoqi Fan, Karttikeya Mangalam, Bo Xiong, Jitendra Malik, and Christoph Feichtenhofer.
\newblock Mvitv2: Improved multiscale vision transformers for classification and detection.
\newblock In {\em Proceedings of the IEEE/CVF Conference on Computer Vision and Pattern Recognition}, pages 4804--4814, 2022.

\bibitem{lin2022frozen}
Ziyi Lin, Shijie Geng, Renrui Zhang, Peng Gao, Gerard De~Melo, Xiaogang Wang, Jifeng Dai, Yu Qiao, and Hongsheng Li.
\newblock Frozen clip models are efficient video learners.
\newblock In {\em European Conference on Computer Vision}, pages 388--404. Springer, 2022.

\bibitem{nishida1999ethogram}
Toshisada Nishida, Takayoshi Kano, Jane Goodall, William~C McGrew, and Michio Nakamura.
\newblock Ethogram and ethnography of mahale chimpanzees.
\newblock {\em Anthropological Science}, 107(2):141--188, 1999.

\bibitem{nishida2010chimpanzee}
Toshisada Nishida, Koichiro Zamma, Takahisa Matsusaka, Agumi Inaba, and William~C McGrew.
\newblock {\em Chimpanzee behavior in the wild: an audio-visual encyclopedia}.
\newblock Springer Science \& Business Media, 2010.

\bibitem{perrett2023}
Toby Perrett, Saptarshi Sinha, Tilo Burghardt, Majid Mirmehdi, and Dima Damen.
\newblock Use your head: Improving long-tail video recognition.
\newblock In {\em Proceedings of the IEEE/CVF Conference on Computer Vision and Pattern Recognition}, pages 2415--2425, 2023.

\bibitem{radford2021learning}
Alec Radford, Jong~Wook Kim, Chris Hallacy, Aditya Ramesh, Gabriel Goh, Sandhini Agarwal, Girish Sastry, Amanda Askell, Pamela Mishkin, Jack Clark, et~al.
\newblock Learning transferable visual models from natural language supervision.
\newblock In {\em International conference on machine learning}, pages 8748--8763. PMLR, 2021.

\bibitem{sakib2020visual}
Faizaan Sakib and Tilo Burghardt.
\newblock Visual recognition of great ape behaviours in the wild.
\newblock In {\em Workshop on the Visual Observation and Analysis of Vertebrate and Insect Behaviour}, 2020.

\bibitem{simonyan2014two}
Karen Simonyan and Andrew Zisserman.
\newblock Two-stream convolutional networks for action recognition in videos.
\newblock In {\em Advances in Neural Information Processing Systems}, volume~27, 2014.

\bibitem{steenweg2017scaling}
Robin Steenweg, Mark Hebblewhite, Roland Kays, Jorge Ahumada, Jason~T Fisher, Cole Burton, Susan~E Townsend, Chris Carbone, J~Marcus Rowcliffe, Jesse Whittington, et~al.
\newblock Scaling-up camera traps: Monitoring the planet's biodiversity with networks of remote sensors.
\newblock {\em Frontiers in Ecology and the Environment}, 15(1):26--34, 2017.

\bibitem{du2017closer}
Du Tran, Heng Wang, Lorenzo Torresani, Jamie Ray, Yann LeCun, and Manohar Paluri.
\newblock A closer look at spatiotemporal convolutions for action recognition.
\newblock In {\em Proceedings of the IEEE Conference on Computer Vision and Pattern Recognition}, pages 6450--6459, 2018.

\bibitem{tuia2022}
Devis Tuia, Benjamin Kellenberger, Sara Beery, Blair~R Costelloe, Silvia Zuffi, Benjamin Risse, Alexander Mathis, Mackenzie~W Mathis, Frank van Langevelde, Tilo Burghardt, et~al.
\newblock Perspectives in machine learning for wildlife conservation.
\newblock {\em Nature communications}, 13(1):1--15, 2022.

\bibitem{wang2022internvideo}
Yi Wang, Kunchang Li, Yizhuo Li, Yinan He, Bingkun Huang, Zhiyu Zhao, Hongjie Zhang, Jilan Xu, Yi Liu, Zun Wang, Sen Xing, Guo Chen, Junting Pan, Jiashuo Yu, Yali Wang, Limin Wang, and Yu Qiao.
\newblock Internvideo: General video foundation models via generative and discriminative learning.
\newblock {\em ArXiv}, abs/2212.03191, 2022.

\bibitem{wang2024exploring}
Yidong Wang, Zhuohao Yu, Jindong Wang, Qiang Heng, Hao Chen, Wei Ye, Rui Xie, Xing Xie, and Shikun Zhang.
\newblock Exploring vision-language models for imbalanced learning.
\newblock {\em International Journal of Computer Vision}, 132(1):224--237, 2024.

\bibitem{xue2022clip}
Hongwei Xue, Yuchong Sun, Bei Liu, Jianlong Fu, Ruihua Song, Houqiang Li, and Jiebo Luo.
\newblock Clip-vip: Adapting pre-trained image-text model to video-language representation alignment.
\newblock {\em ArXiv}, abs/2209.06430, 2022.

\bibitem{yu2022coca}
Jiahui Yu, Zirui Wang, Vijay Vasudevan, Legg Yeung, Mojtaba Seyedhosseini, and Yonghui Wu.
\newblock Coca: Contrastive captioners are image-text foundation models.
\newblock {\em Trans. Mach. Learn. Res.}, 2022, 2022.

\bibitem{zamma_matsusaka_2015}
Koichiro Zamma and Takahisa Matsusaka.
\newblock {\em Ethograms and the diversity of behaviors}, page 510–518.
\newblock Cambridge University Press, 2015.

\bibitem{zhang2024vision}
Jingyi Zhang, Jiaxing Huang, Sheng Jin, and Shijian Lu.
\newblock Vision-language models for vision tasks: A survey.
\newblock {\em IEEE Transactions on Pattern Analysis and Machine Intelligence}, 2024.

\bibitem{zhao2023learning}
Yue Zhao, Ishan Misra, Philipp Kr{\"a}henb{\"u}hl, and Rohit Girdhar.
\newblock Learning video representations from large language models.
\newblock In {\em Proceedings of the IEEE/CVF Conference on Computer Vision and Pattern Recognition}, pages 6586--6597, 2023.

\end{thebibliography}
}

\end{document}